\documentclass[conference]{IEEEtran}
\IEEEoverridecommandlockouts
\usepackage{cite}
\usepackage{amsmath,amssymb,amsfonts}
\usepackage{algorithmic}
\usepackage{graphicx}
\usepackage{textcomp}
\usepackage{xcolor}
\usepackage{adjustbox}
\usepackage{xspace}
\usepackage{bbding}
\usepackage{authblk}
\usepackage{multirow}
\usepackage[pagebackref=true,breaklinks=true,colorlinks,bookmarks=false]{hyperref}

\makeatletter
\DeclareRobustCommand\onedot{\futurelet\@let@token\@onedot}
\def\@onedot{\ifx\@let@token.\else.\null\fi\xspace}

\def\etal{\emph{et al}\onedot}
\makeatother

\def\BibTeX{{\rm B\kern-.05em{\sc i\kern-.025em b}\kern-.08em
    T\kern-.1667em\lower.7ex\hbox{E}\kern-.125emX}}
\begin{document}

\title{ShaDocFormer: A Shadow-Attentive Threshold Detector With Cascaded Fusion Refiner for Document Shadow Removal
}

\author[1\IEEEauthorrefmark{1}]{Weiwen Chen}
\author[1\IEEEauthorrefmark{1}]{Yingtie Lei\thanks{\IEEEauthorrefmark{1} Equal contributions}}
\author[1]{Shenghong Luo}
\author[2]{Ziyang Zhou}
\author[2]{Mingxian Li}
\author[1\IEEEauthorrefmark{2}]{Chi-Man Pun\thanks{\IEEEauthorrefmark{2} Corresponding author}}
\affil[1]{University of Macau, Macau, Macao}
\affil[2]{Huizhou University, Huizhou, China}

\maketitle

\begin{abstract}
Document shadow is a common issue that arises when capturing documents using mobile devices, which significantly impacts readability. 
Current methods encounter various challenges, including inaccurate detection of shadow masks and estimation of illumination. 
In this paper, we propose ShaDocFormer, a Transformer-based architecture that integrates traditional methodologies and deep learning techniques to tackle the problem of document shadow removal. 
The ShaDocFormer architecture comprises two components: the Shadow-attentive Threshold Detector (STD) and the Cascaded Fusion Refiner (CFR). 
The STD module employs a traditional thresholding technique and leverages the attention mechanism of the Transformer to gather global information, thereby enabling precise detection of shadow masks. 
The cascaded and aggregative structure of the CFR module facilitates a coarse-to-fine restoration process for the entire image. 
As a result, ShaDocFormer excels in accurately detecting and capturing variations in both shadow and illumination, thereby enabling effective removal of shadows. 
Extensive experiments demonstrate that ShaDocFormer outperforms current state-of-the-art methods in both qualitative and quantitative measurements.
\end{abstract}


\section{Introduction}
Over the past few years, the technology behind smartphone cameras has made considerable progress, leading to their widespread adoption for digitizing text from paper documents. Despite these advancements, a common issue persists with the presence of shadows in the photographed document images. These shadows can cause problems in discerning text, as they often cast areas of darkness that can be close in color to the actual text, making it hard to distinguish the characters. This issue poses a significant obstacle in the readability and subsequent processing of document images, as it interferes with the accurate extraction of textual information.

\begin{figure}[t]
    \begin{minipage}[b]{1.0\linewidth}
        \begin{minipage}[b]{.32\linewidth}
            \centering
            \centerline{\includegraphics[width=\linewidth]{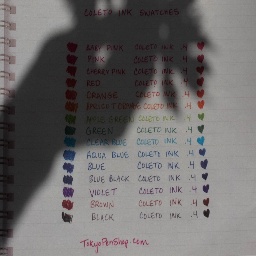}}
            \centerline{(a) Input}\medskip
        \end{minipage}
        \hfill
        \begin{minipage}[b]{.32\linewidth}
            \centering
            \centerline{\includegraphics[width=\linewidth]{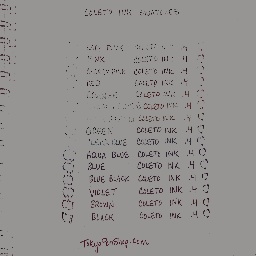}}
            \centerline{(b) Liu~\etal}\medskip
        \end{minipage}
        \hfill
        \begin{minipage}[b]{0.32\linewidth}
            \centering
            \centerline{\includegraphics[width=\linewidth]{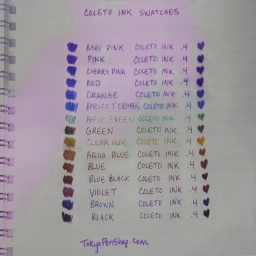}}
            \centerline{(c) BEDSR-Net}\medskip
        \end{minipage}
    \end{minipage}
    \begin{minipage}[b]{1.0\linewidth}
        \begin{minipage}[b]{.32\linewidth}
            \centering
            \centerline{\includegraphics[width=\linewidth]{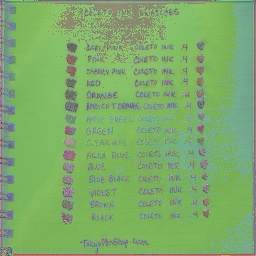}}
            \centerline{(d) BGShadowNet}\medskip
        \end{minipage}
        \hfill
        \begin{minipage}[b]{.32\linewidth}
            \centering
            \centerline{\includegraphics[width=\linewidth]{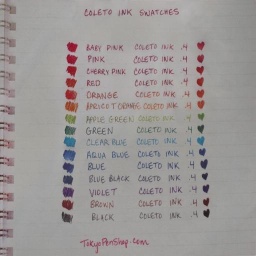}}
            \centerline{(e) Ours}\medskip
        \end{minipage}
        \hfill
        \begin{minipage}[b]{0.32\linewidth}
            \centering
            \centerline{\includegraphics[width=\linewidth]{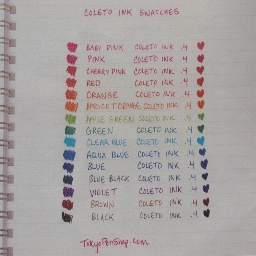}}
            \centerline{(f) Target}\medskip
        \end{minipage}
    \end{minipage}
    \caption{The visual results of the input document shadow image (a), along with the results of the traditional method (b), learning-based methods (c) and (d), ours (e), and the target (f). Our model effectively removes the shadow while preserving the content and color information of the original document.	
    } 
    \label{fig:teaser}
\end{figure}

A range of techniques, including classic strategies and machine learning-based models, have been explored to mitigate the impact of shadows in images of documents. Traditional approaches typically focus on analyzing changes in brightness or assessing patterns of light in scanned documents through rule-based algorithms. Although these methods can be effective in controlled conditions, they tend to fall short in practical applications. These shortcomings are evident in scenarios with complex lighting, as can be seen in Fig.~\ref{fig:teaser} (b), where traditional methods struggle to accurately distinguish between shadows and text, leading to less than satisfactory results in real-life situations.

In recent years, a myriad of models have been proposed to remove shadows from either natural images or document imagery~\cite{wang2018stacked,lin2020bedsr,fu2021auto,jin2021dc,zhang2023document}. However, certain models~\cite{wang2018stacked,fu2021auto} necessitate binary shadow masks during the training phase, whereas some models require a two-stage training process~\cite{lin2020bedsr,zhang2023document}, which may lead to imprecise estimations of the background, as shown in Fig.~\ref{fig:teaser} (c) and (d).

To address the aforementioned issues, we propose ShaDocFormer, a Transformer-based method comprising two modules: the Shadow-attentive Threshold Detector (STD) and the Cascaded Fusion Refiner (CFR).
In the STD module, we integrate the transformer with the Otsu algorithm~\cite{otsu1979threshold}, which can effectively detect and segment shadows by adaptively selecting thresholds. By combining the Otsu algorithm with the Vision Transformer~\cite{dosovitskiy2020vit,li2022wavenhancer,Li_2023_ICCV,ijcai2023p129,luo2023devignet,jing2023ta}, we enable the extraction of features and capture of global information, thereby reducing computational burden while ensuring accurate shadow masks.
Inspired by~\cite{chen2022simple,Chen2024-dg,huang2023mr,10365931,liu2023explicit,liu2023coordfill,liu2024depth,li2023cee,li2022monocular}, the CFR module facilitates effective restoration of shadow areas by utilizing masks to guide the restoration process. The cascaded and aggregative structures employed by CFR enable precise restoration of shadow areas while preserving the overall structure and features of the image, as shown in Fig.~\ref{fig:teaser} (e).

We summarize the main contribution as follows:

\begin{enumerate}
    \item We propose ShaDocFormer, a Transformer-based network focusing on recovering document information in shadow areas. Qualitative and quantitative experiments have demonstrated the superiority of our method over state-of-the-art approaches.
    \item We propose Shadow-attentive Threshold Detector (STD) module, a lightweight module that combines a traditional algorithm with the Vision Transformer to efficiently extract shadow masks.
    \item We propose Cascaded Fusion Refiner (CFR) module that accurately recovers document information in the shadow regions.
\end{enumerate}

\begin{figure*}[ht]
\begin{minipage}[b]{1.0\linewidth}
    \includegraphics[width=\linewidth]{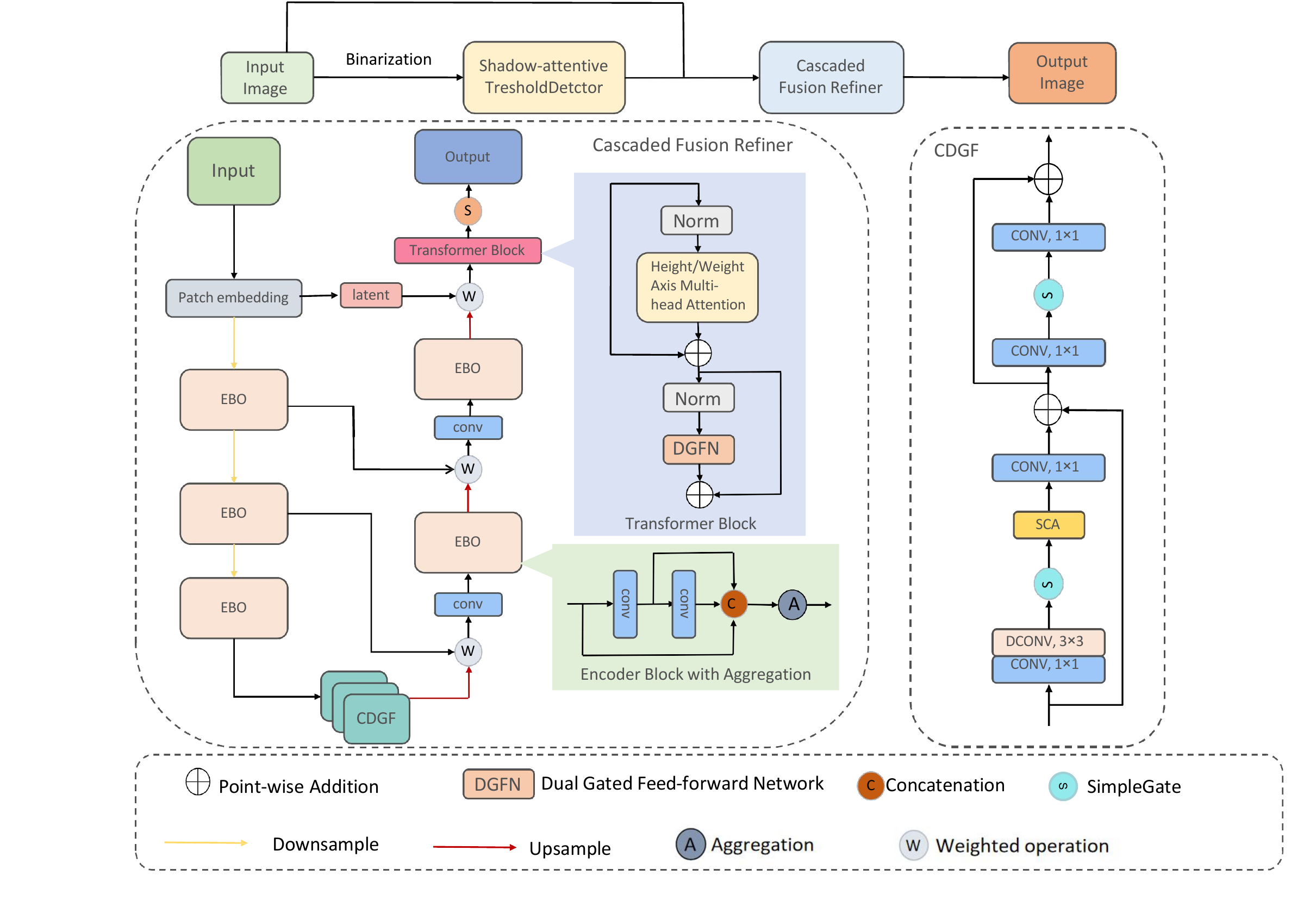}
\end{minipage}
\caption{Model diagram of the proposed Cascaded Fusion Refiner framework and the Convolutional Depth
Grouped Fusion Net framework.}
\label{fig:model}
\end{figure*}

\section{Related Work}
\subsection{Traditional Methods}
In the realm of document shadow removal, the norm has been to conduct detailed evaluations of reflectance and light intensity in shadow-laden areas. This established technique is focused on detecting and reviving the hidden true colors and brightness levels, with the goal of restoring the document to its original visual state~\cite{shah2018iterative,wang2019effective,wang2020shadow,liu2023shadow}. The aim is to preserve not just the document's aesthetics but also its readability and the authenticity of its information.

Complementing these techniques, the research community has introduced specialized yet small-scale datasets designed for precise evaluation purposes~\cite{bako2017removing,kligler2018document,jung2018water}. Despite their limited size, these datasets are pivotal, providing a focused means for rigorously testing and benchmarking the performance of shadow removal algorithms.

\subsection{Learning-based Methods}
DeShadowNet~\cite{qu2017deshadownet} is a deep learning-based technique that effectively removes shadows from images by combining CNNs and Conditional Random Fields (CRFs). ST-CGAN~\cite{wang2018stacked} consists of two consecutive conditional generative adversarial networks (GAN)~\cite{Zuo2023BrainFN,jing2024estimating,zuo2023diffgan}, where the first network is specifically designed for shadow detection and the second network is focused on shadow removal. 
Mask-ShadowGAN~\cite{hu2019mask} simultaneously learns to produce shadow masks and learns to remove shadows to maximize the overall performance.
BEDSR-Net~\cite{lin2020bedsr} is a neural network exclusively tailored for shadow removal in document images. It comprises the Background Estimation Network and a dedicated shadow removal module.  DHAN~\cite{cun2020towards} is a deep learning method for image dehazing, which extracts information about haze from the input image and uses back-propagation to restore the clarity and details of the image. LG-ShadowNet~\cite{liu2021shadow} is a deep learning-based framework proposed for shadow detection and removal in images. The framework consists of two main components: the ShadowNet and the StructureNet. The ShadowNet is responsible for shadow detection, while the StructureNet focuses on shadow removal. AEFNet~\cite{fu2021auto} achieves color alignment in shadow areas by estimating the over-exposed image condition and integrating it with the original input in the shadow region. DC-ShadowNet~\cite{jin2021dc} is an unsupervised domain classifier-guided network specifically developed for single image shadow removal. Unfolding~\cite{zhu2022efficient} offers an efficient and effective solution to the problem of blind deblurring, leveraging deep learning techniques to enhance the traditional alternating optimization approach. Zhang \etal~\cite{zhang2023document} propose the RDD dataset, a large-scale real-world dataset specifically designed for document shadow removal. Additionally, they present a background-guided document image shadow removal network. ShaDocNet~\cite{chen2023shadocnet} combines convolutional neural networks (CNNs) and long short-term memory (LSTM) networks. This hybrid model extracts rich semantic features from raw text data and utilizes these features for document classification. DMTN~\cite{dmtn} is a deep learning model used for image-to-image translation tasks. It utilizes a mutual transformation network that consists of two main components: the image transformation network and the feature transformation network. TBRNet~\cite{tbr} is a deep learning model designed to tackle the problem of text-image matching. It aims to learn the correlation between textual descriptions and image content to facilitate tasks such as image retrieval, caption generation, and cross-modal retrieval.

\section{Methodology}

\subsection{Overview}
\begin{figure}[t]
\begin{minipage}[b]{1.0\linewidth}
    \includegraphics[width=\linewidth]{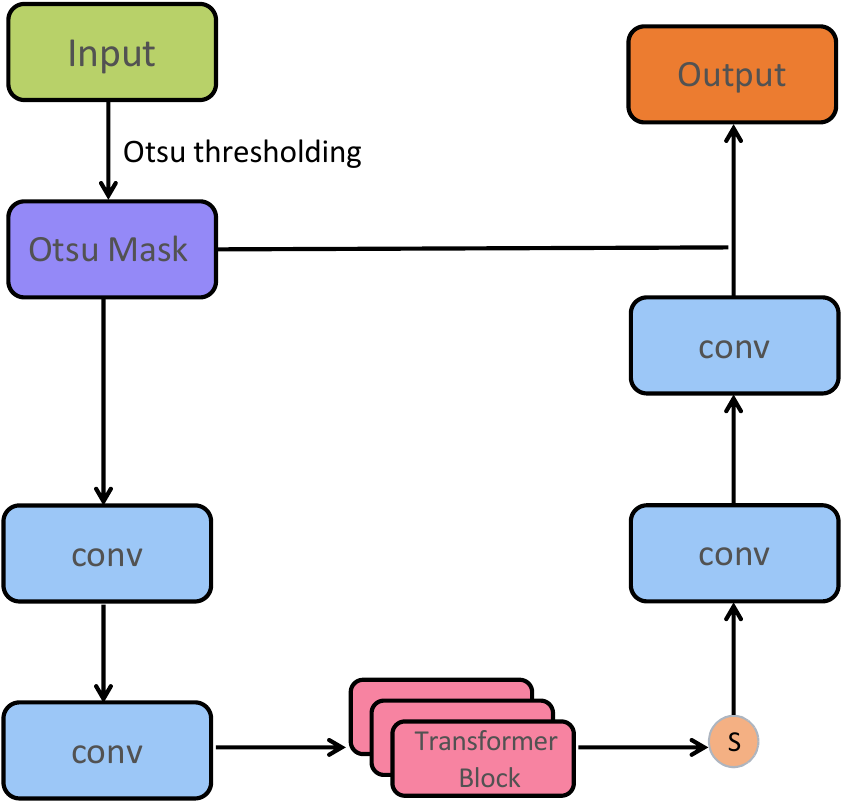}
\end{minipage}
\caption{Model diagram of the proposed Shadow-attentive Threshold Detector framework. It combines
the intricacies of convolutional layers with the contextual prowess of Transformer blocks for enhanced image processing.}
\label{fig:SATD}
\end{figure}

The process of accurately detecting shadow masks and estimating the underlying illumination conditions plays a pivotal role in the domain of document analysis, particularly when the task at hand involves the intricate challenge of document shadow removal. Shadows cast on document images can significantly degrade the legibility and visual quality of the textual content, thereby impeding subsequent processes such as document digitization and optical character recognition (OCR). To effectively mitigate this issue, we introduce a novel method, designated as the ShaDocFormer. When a document image with shadows is given as input, the Shadow-attentive Threshold Detector (STD) leverages advanced attention mechanisms to identify the shadow masks~\cite{zhou2024qean}. Subsequently, the Cascaded Fusion Refiner (CFR) restores the document information within the shaded regions. To illustrate the comprehensive architecture and the operational flow of our model, we have encapsulated its essence in a graphical representation. Fig.~\ref{fig:model} presents the overall diagram of our model. Fig.~\ref{fig:SATD} presents the structure of Shadow-attentive Threshold Detector. These figures serve as a kind of visual aid to convey the methodological structure of the proposed model and to provide an intuitive understanding of the sequential procedures involved in the shadow removal process.

\subsection{Shadow-attentive Threshold Detector}
We propose a novel Shadow-attentive Threshold Detector (STD), a neural network architecture that adeptly combines the intricacies of convolutional layers with the contextual prowess of Transformer blocks for enhanced image processing~\cite{chen2023medprompt,li2023optimized}. The STD model, drawing inspiration from~\cite{otsu1979threshold}, particularly leverages the Otsu algorithm to automate the thresholding process. This strategic choice ensures the optimal segregation of shadow regions by minimizing within-class variance and amplifying between-class variance, thus laying a robust foundation for shadow detection.

Upon this preliminary segmentation, the STD employs a series of convolutional layers to meticulously extract hierarchical features from the binarized images. The integration of activation functions introduces the necessary non-linearity to capture complex patterns within the data. Complementing this, the Transformer blocks within the STD model apply their attention mechanisms to distill and interpret the global contextual information from the document images, providing a nuanced understanding of the shadowed areas.

The interaction between the convolutional layers and Transformer blocks in STD allows for a sophisticated extraction of deep image features from the binary mask, thereby enhancing the accuracy of shadow detection. This approach encapsulates a blend of classical image thresholding technique and advanced neural network design, positioning the STD as a cutting-edge solution in the field of document image analysis.

\subsection{Cascaded Fusion Refiner}
We propose Cascaded Fusion Refiner (CFR), a paradigm engineered to systematically undertake the encoding, decoding, and refinement of image data. At the outset, the proposed CFR paradigm embarks on the task of dissecting the input image into a mosaic of overlapping image segments, each of which is subsequently subjected to a meticulous encoding process. This intricate operation is the precursor to a structured downscaling process, wherein the image undergoes a trifold succession of downsampling phases. Such a process is instrumental in meticulously molding the input image into a U-shaped configuration, a transformative step that not only diminishes the spatial dimensions of the image but also efficaciously amplifies the depth of the input image features through a series of convolutive manipulations.

In order to enhance the fidelity of the image representation, our model harnesses the potential of Convolutional Depth Grouped Fusion Net (CDGF). This innovative module is adept in the art of feature extraction and augmentation, employing a multi-faceted approach that encompasses convolutions, normalization, dynamic scaling, and the integration of residual concatenations. Our CDGF module, drawing inspiration from seminal work ~\cite{chen2022simple}, is proficient in apprehending a comprehensive spectrum of shadow features, which is achieved by the integration of a dedicated Transformer block~\cite{wang2023ultra}.

The CDGF model is further augmented by the incorporation of a Multi-head Attention mechanism and a Dual Gated Feed-Forward Network (DGFN). These components collectively function as the cognitive center of the model, playing a pivotal role in not only curtailing the computational overhead but also in capturing an expanded repertoire of image features. This dual-pronged approach ensures a refined feature set that is both comprehensive and computationally efficient.

Subsequent to the extraction and enhancement of features, CFR embarks on an upscaling journey, meticulously restoring the salient details of the image by inversely amplifying the dimensions of the input image features through a series of convolution operations. This is succeeded by the application of a Spatial Pyramid Pooling (SPP) operation to the encoded features, executed by an adept SPP module, which serves to amalgamate features at varying scales and dimensions.

During the final stage of the Composite Feature Refinement (CFR) process, the Transformer block emerges as a key component, orchestrating the final enhancement of the features. Within this advanced module, the input features are meticulously fine-tuned, allowing the Transformer to act as a forge that tempers and refines the image data. This process is designed to amplify the clarity, enrich the texture details, and elevate the overall quality of the final image output. As the features pass through the Transformer block, they are not simply altered but are fundamentally transformed, optimizing the image for better visual interpretation. The CFR framework, through this elaborate process, aims to redefine excellence in image processing, proposing an innovative approach that is both sturdy and effective. By leveraging this refined methodology, the CFR framework endeavors to consistently produce images that are not only visually appealing but also demonstrate a marked improvement in the representation of intricate details and contrast. This commitment to enhancing image quality positions the CFR framework at the forefront of technological advancements in image processing, setting a new standard for accuracy and detail in the field.

\subsection{Objective Function}
In this study, the objective function is formulated as a combination of the mean square error loss ($\mathcal{L}_{MSE}$), structural similarity loss ($\mathcal{L}_{SSIM}$)~\cite{wang2004image}, and perceptual loss ($\mathcal{L}_P$)~\cite{zhang2018unreasonable}. The total loss function can be represented as follows:	
\begin{equation}
\mathcal{L}_{total} = \mathcal{L}_{MSE} + 0.3\times \mathcal{L}_{SSIM} + 0.7\times\mathcal{L}_{P},
\end{equation}

The Mean Square Error (MSE) Loss is defined as:
\begin{equation}
\mathcal{L}_{MSE} = \frac{1}{N} \sum_{i=1}^{N} (I_i - K_i)^2
\end{equation}
where $I$ represents the reference image, $K$ indicates the reconstructed or predicted image; $N$ is the number of pixels in each image.

The Structural Similarity (SSIM) Index is defined as:
\begin{equation}
\text{SSIM}(x, y) = \frac{(2\mu_x \mu_y + c_1)(2\sigma_{xy} + c_2)}{(\mu_x^2 + \mu_y^2 + c_1)(\sigma_x^2 + \sigma_y^2 + c_2)}
\end{equation}
where $\mu_x$, and $\mu_y$ are the averages of $x$ and $y$; $\sigma_x^2$, $\sigma_y^2$ represent the variances of $x$, $y$, $\sigma_{xy}$ indicates the covariance of $x$ and $y$; $c_1 = (k_1 L)^2$, and $c_2 = (k_2 L)^2$ are two constants to stabilize the division with a weak denominator; $L$ symbolizes the dynamic range of the pixel-values (typically this is $2^{\text{bits per pixel}} - 1$); $k_1 = 0.01$, and $k_2 = 0.03$ are small constants by default.

The SSIM loss used in optimization problems is then:
\begin{equation}
\mathcal{L}_{SSIM} = 1 - \text{SSIM}(I, K)
\end{equation}

The Perceptual Loss between a reconstructed image $\hat{y}$ and a target image $y$ can be defined as:
\begin{equation}
\mathcal{L}_{P} = \sum_{l=1}^{L} \frac{1}{N_l} \sum_{i=1}^{H_l} \sum_{j=1}^{W_l} \left( \phi_l(\hat{y})_{i,j} - \phi_l(y)_{i,j} \right)^2
\end{equation}
where $\phi_l(\cdot)$ denotes the feature map obtained by a particular layer $l$ within a pre-trained CNN; $L$ is the number of layers chosen from the CNN to compute the loss; The dimensions of the feature maps for layer $l$ are given by $H_l$ and $W_l$, which represent height and width, respectively; $N_l$ represents the number of elements in the feature map of layer $l$, and is calculated as the product of $C_l \times H_l \times W_l$, where $C_l$ is the number of channels for that particular layer.

\begin{figure*}[ht]
    \centering
    \begin{minipage}[b]{0.9\linewidth}
        \begin{minipage}[b]{.16\linewidth}
            \centering
            \centerline{\includegraphics[width=\linewidth]{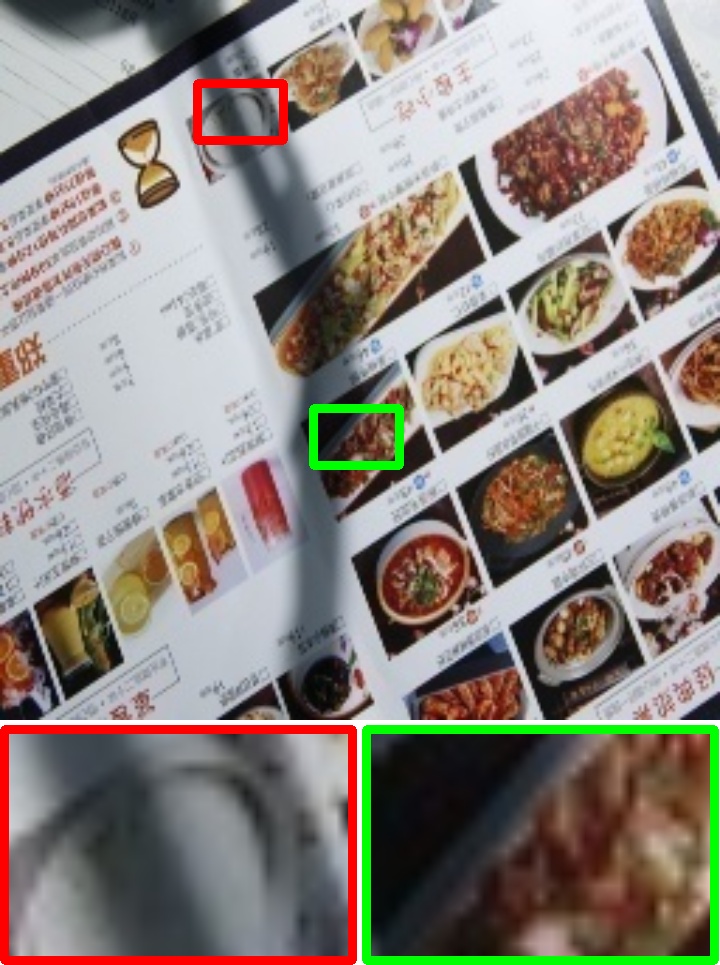}}
        \end{minipage}
        \hfill
        \begin{minipage}[b]{.16\linewidth}
            \centering
            \centerline{\includegraphics[width=\linewidth]{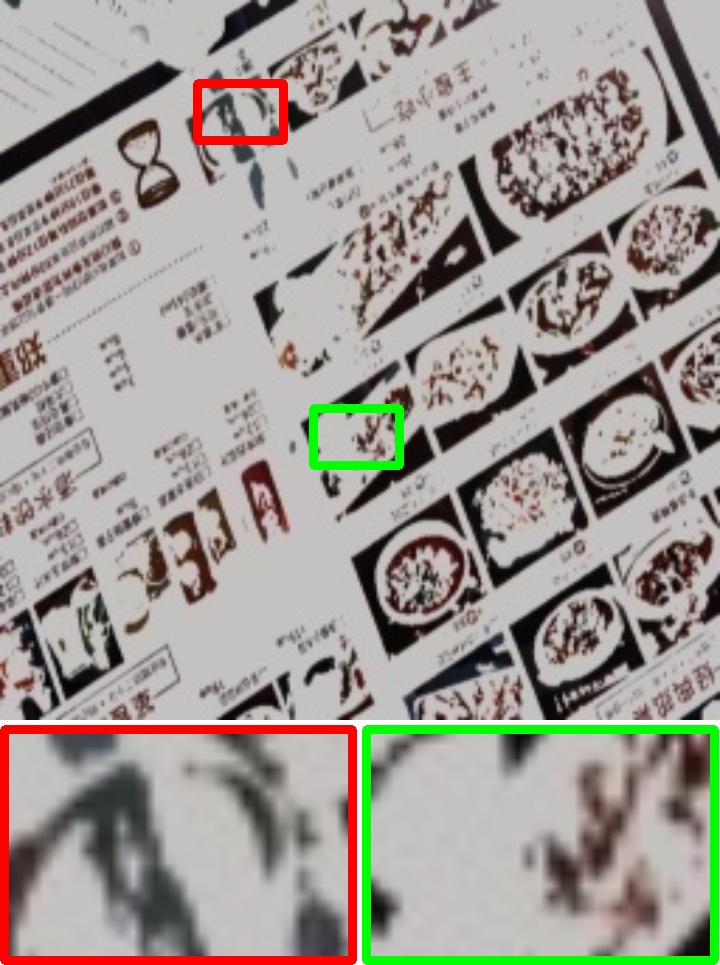}}
        \end{minipage}
        \hfill
        \begin{minipage}[b]{0.16\linewidth}
            \centering
            \centerline{\includegraphics[width=\linewidth]{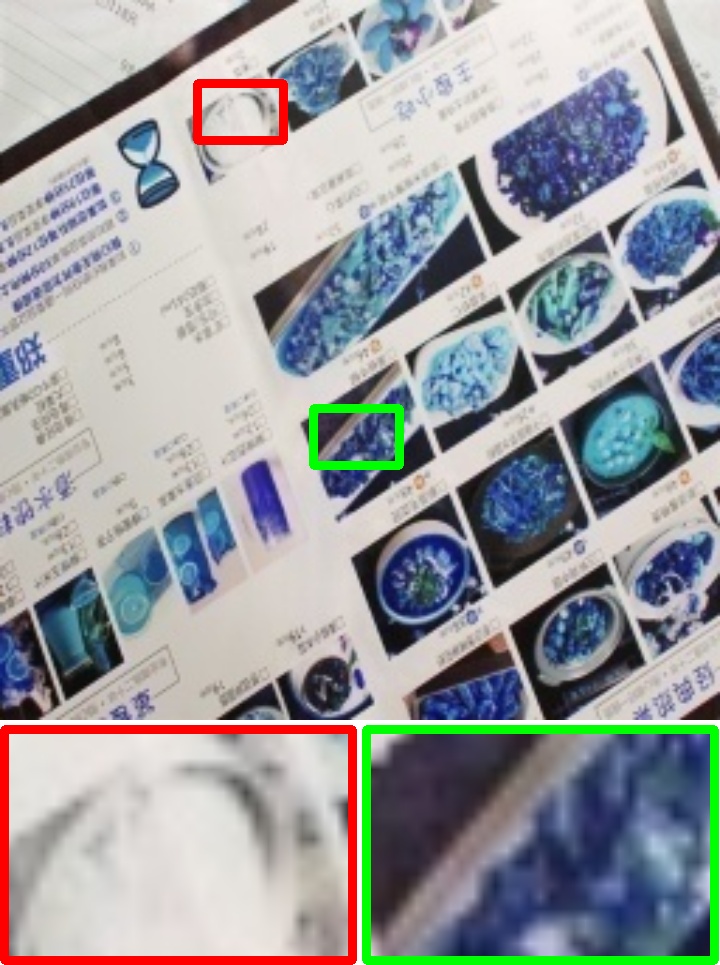}}
        \end{minipage}
        \hfill
        \begin{minipage}[b]{.16\linewidth}
            \centering
            \centerline{\includegraphics[width=\linewidth]{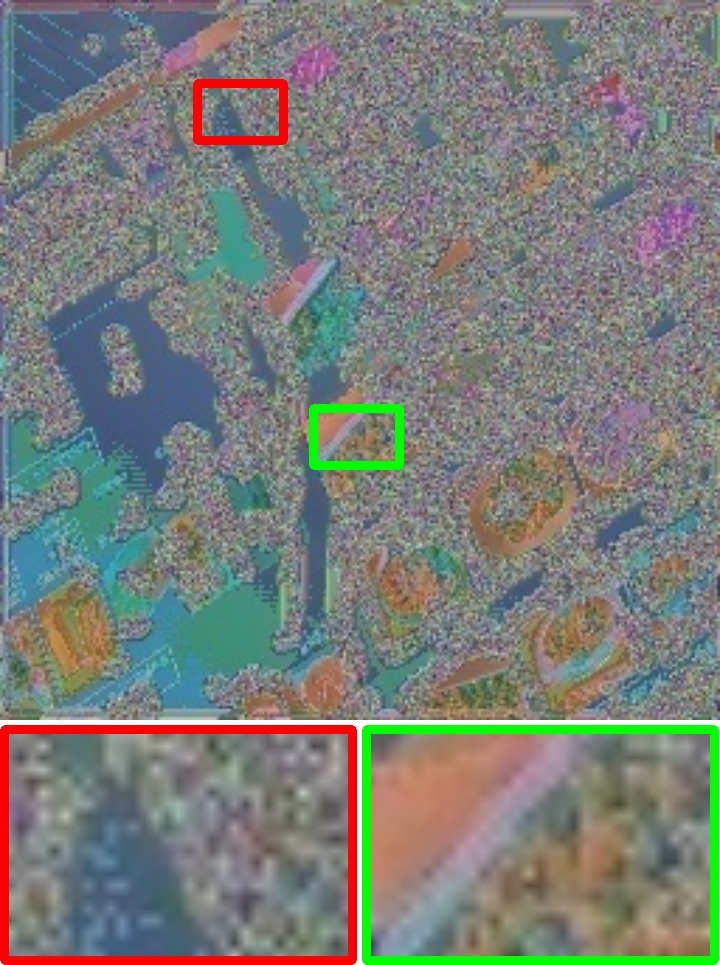}}
        \end{minipage}
        \hfill
        \begin{minipage}[b]{.16\linewidth}
            \centering
            \centerline{\includegraphics[width=\linewidth]{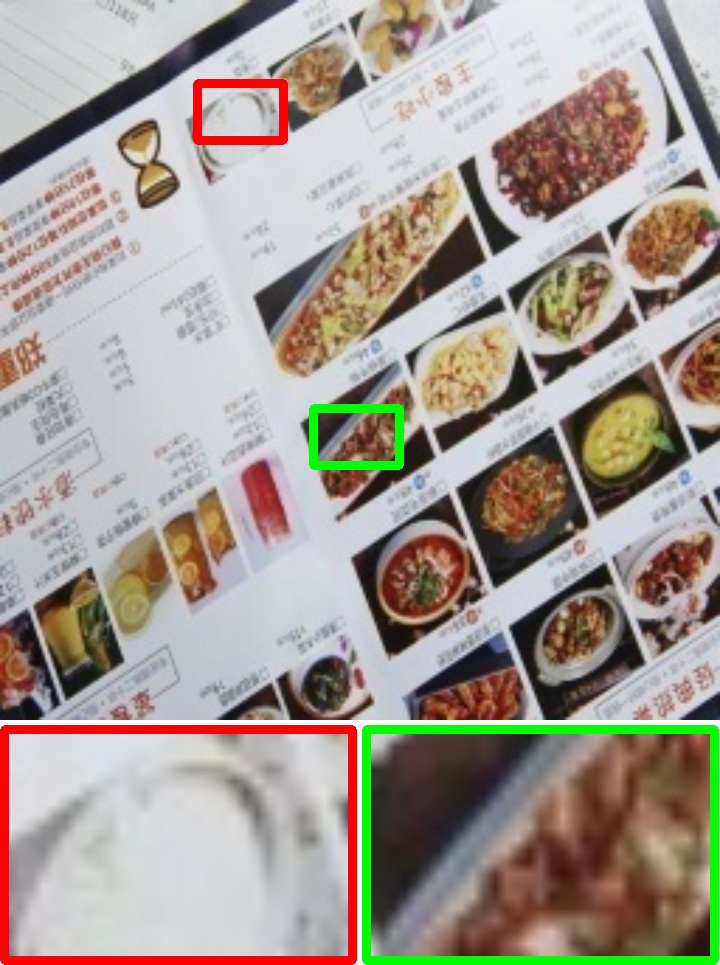}}
        \end{minipage}
        \hfill
        \begin{minipage}[b]{0.16\linewidth}
            \centering
            \centerline{\includegraphics[width=\linewidth]{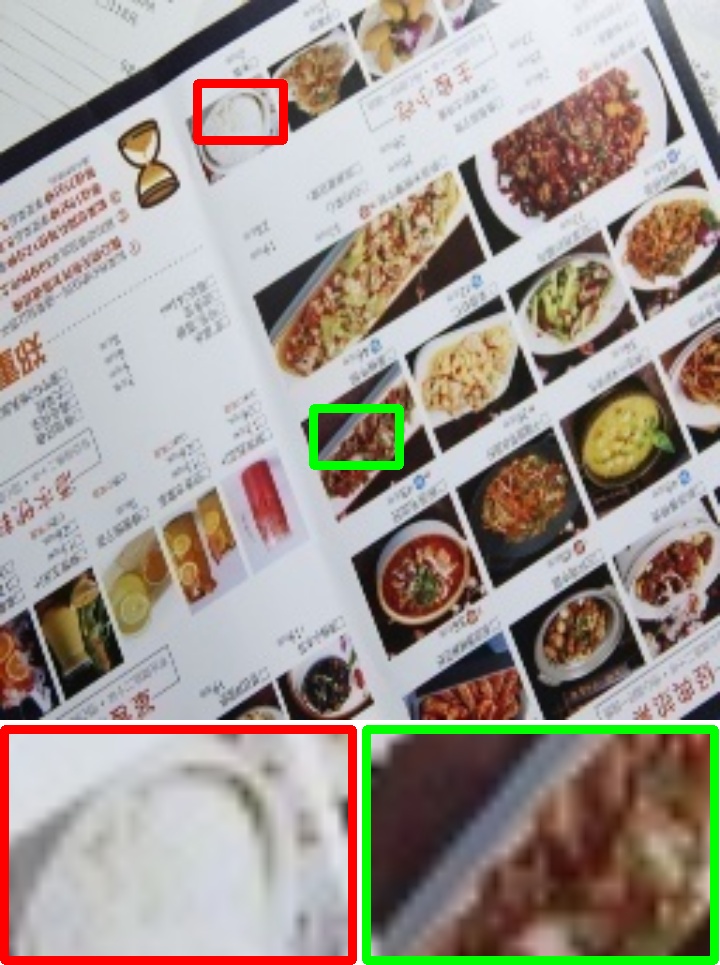}}
        \end{minipage}
    \end{minipage}

    \begin{minipage}[b]{0.9\linewidth}
        \begin{minipage}[b]{.16\linewidth}
            \centering
            \centerline{\includegraphics[width=\linewidth]{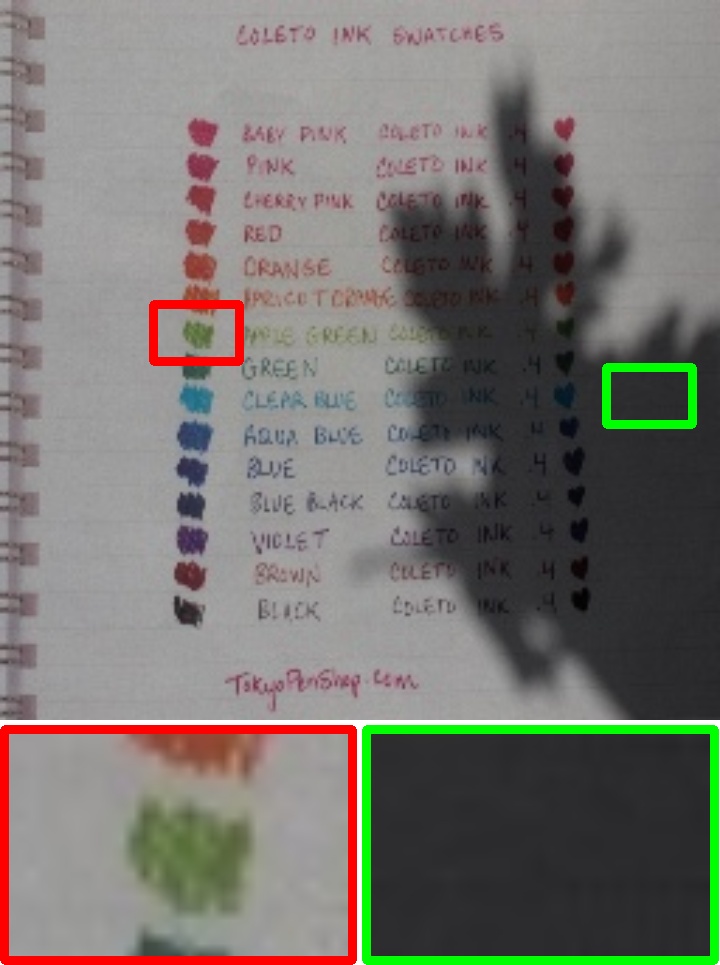}}
        \end{minipage}
        \hfill
        \begin{minipage}[b]{.16\linewidth}
            \centering
            \centerline{\includegraphics[width=\linewidth]{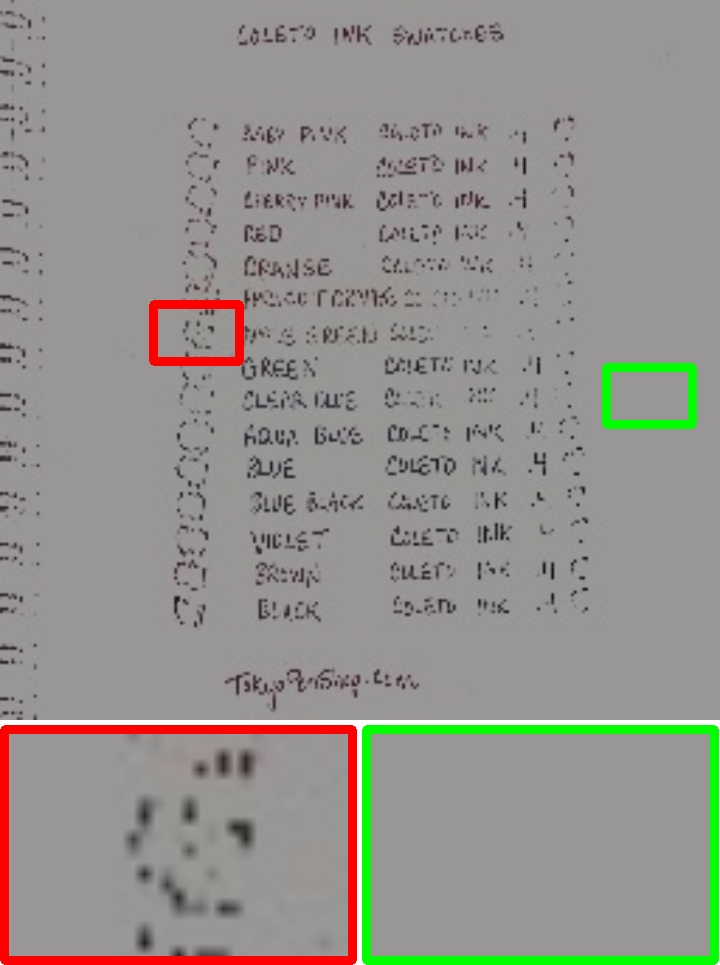}}
        \end{minipage}
        \hfill
        \begin{minipage}[b]{0.16\linewidth}
            \centering
            \centerline{\includegraphics[width=\linewidth]{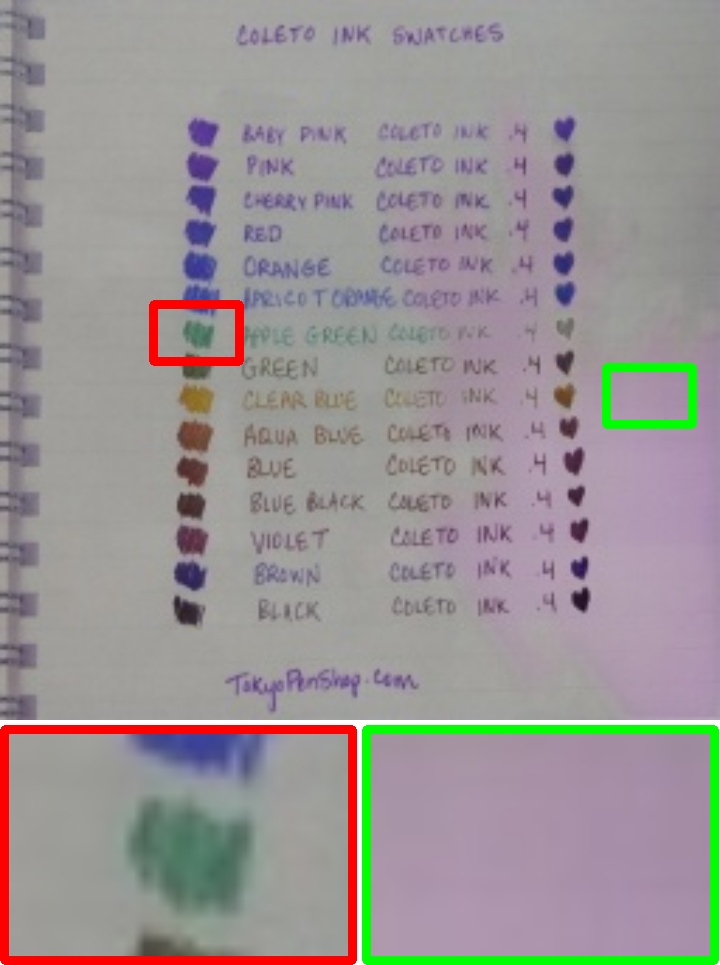}}
        \end{minipage}
        \hfill
        \begin{minipage}[b]{.16\linewidth}
            \centering
            \centerline{\includegraphics[width=\linewidth]{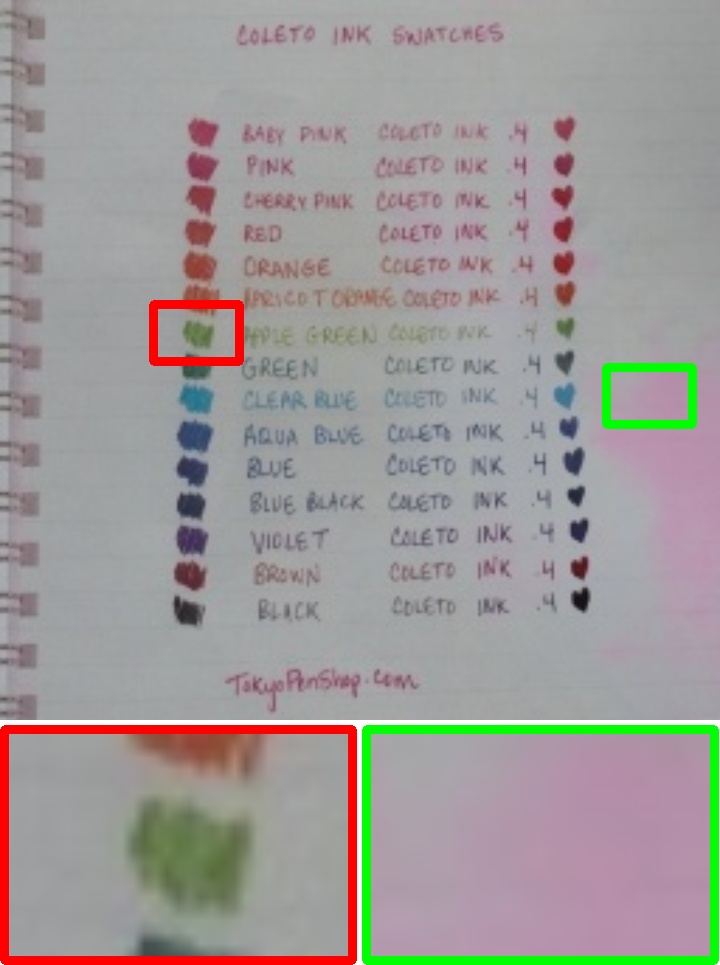}}
        \end{minipage}
        \hfill
        \begin{minipage}[b]{.16\linewidth}
            \centering
            \centerline{\includegraphics[width=\linewidth]{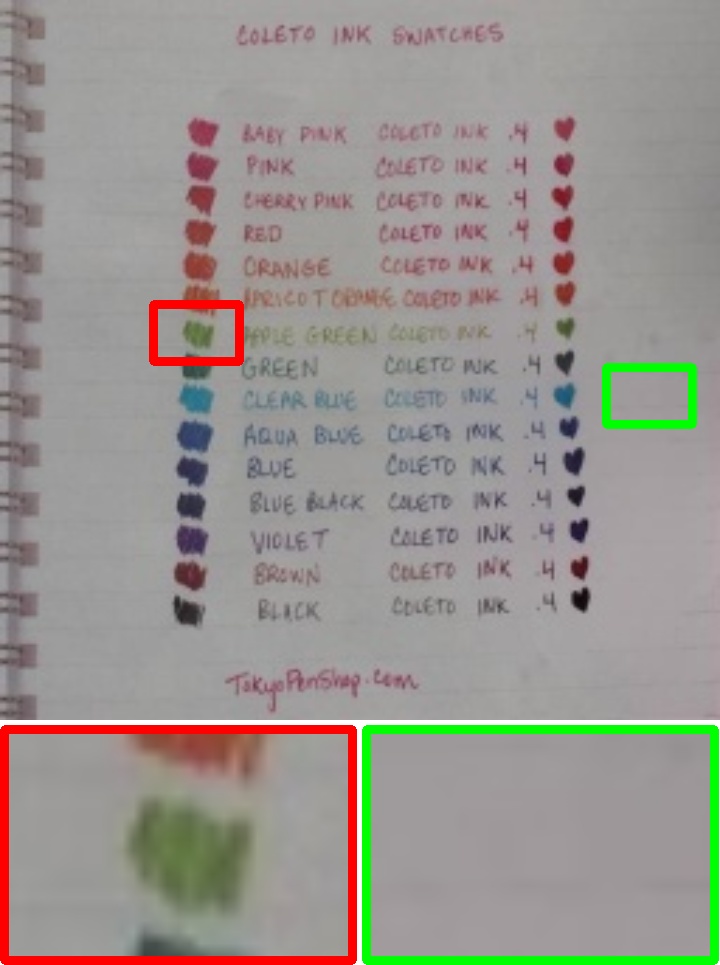}}
        \end{minipage}
        \hfill
        \begin{minipage}[b]{0.16\linewidth}
            \centering
            \centerline{\includegraphics[width=\linewidth]{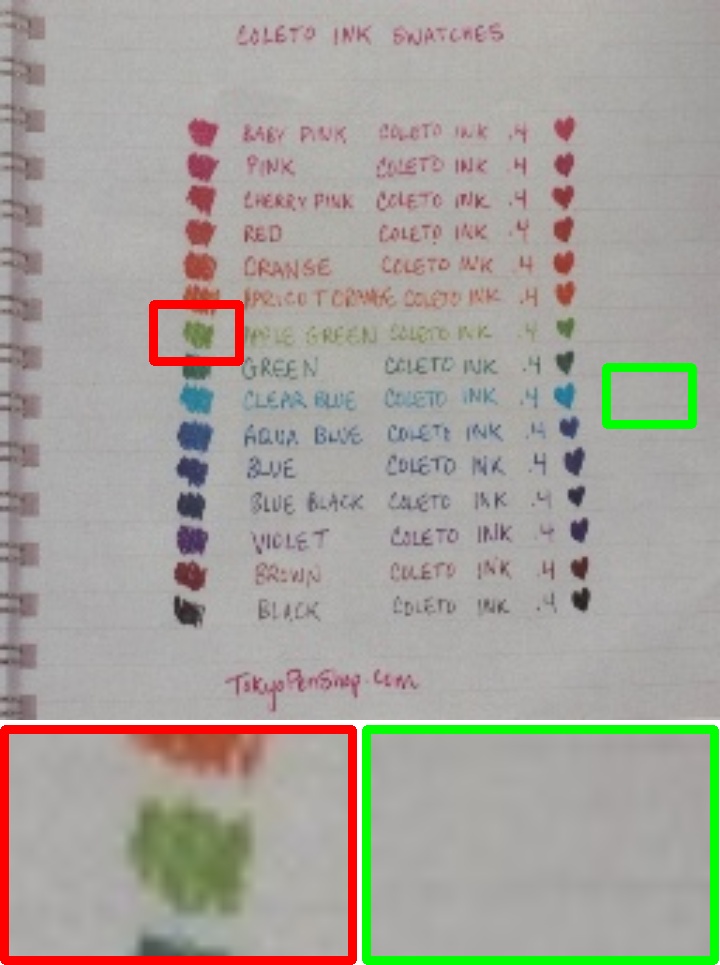}}
        \end{minipage}
    \end{minipage}
    \begin{minipage}[b]{0.9\linewidth}
        \begin{minipage}[b]{.16\linewidth}
            \centering
            \centerline{\includegraphics[width=\linewidth]{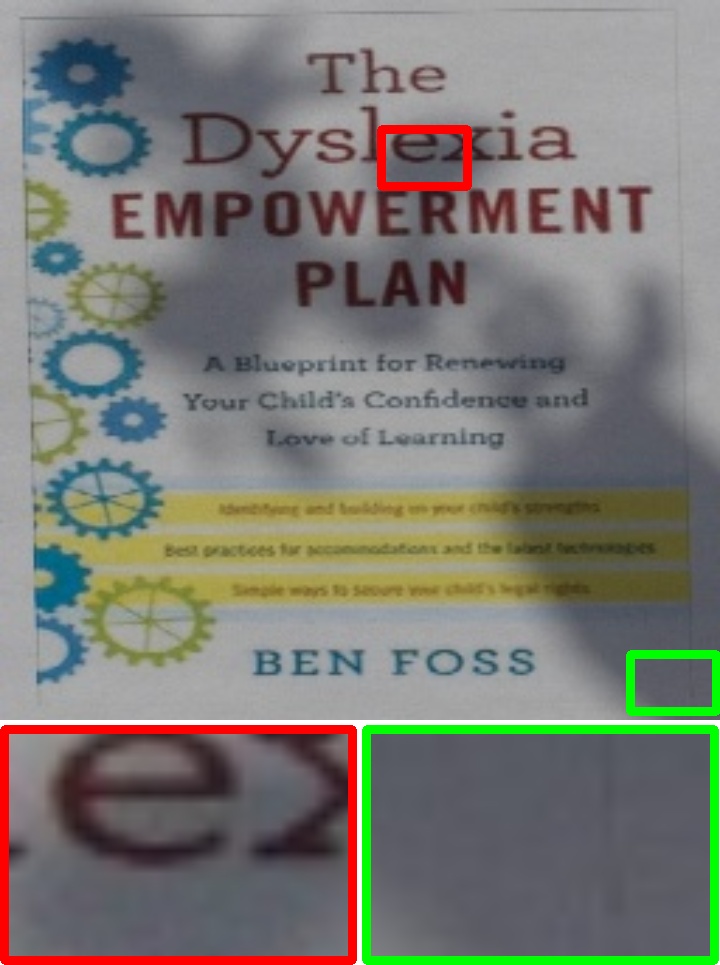}}
        \end{minipage}
        \hfill
        \begin{minipage}[b]{.16\linewidth}
            \centering
            \centerline{\includegraphics[width=\linewidth]{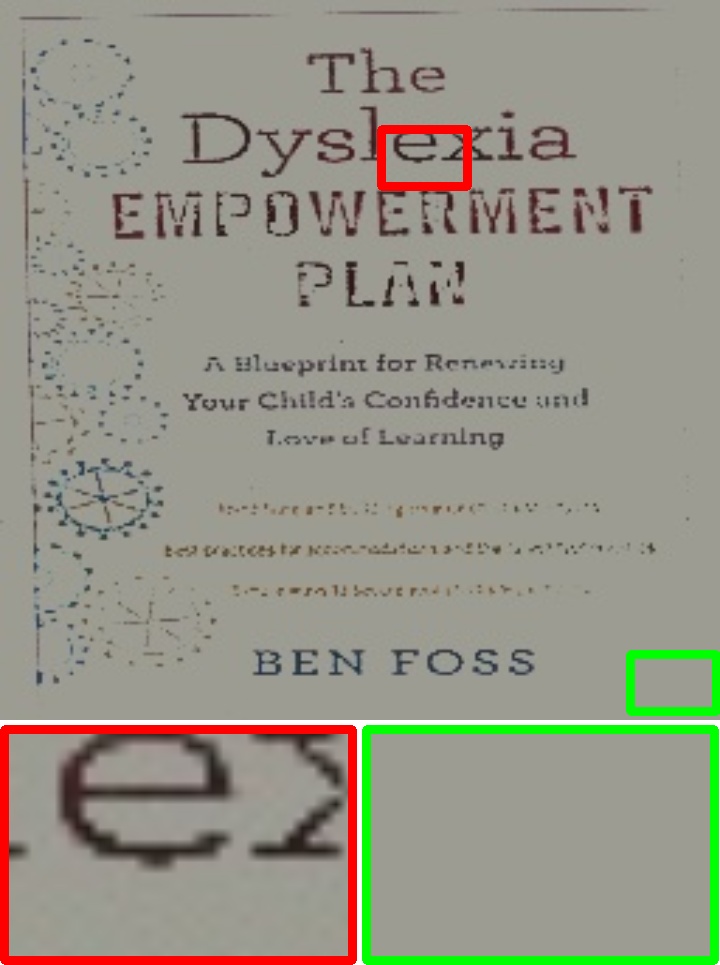}}
        \end{minipage}
        \hfill
        \begin{minipage}[b]{0.16\linewidth}
            \centering
            \centerline{\includegraphics[width=\linewidth]{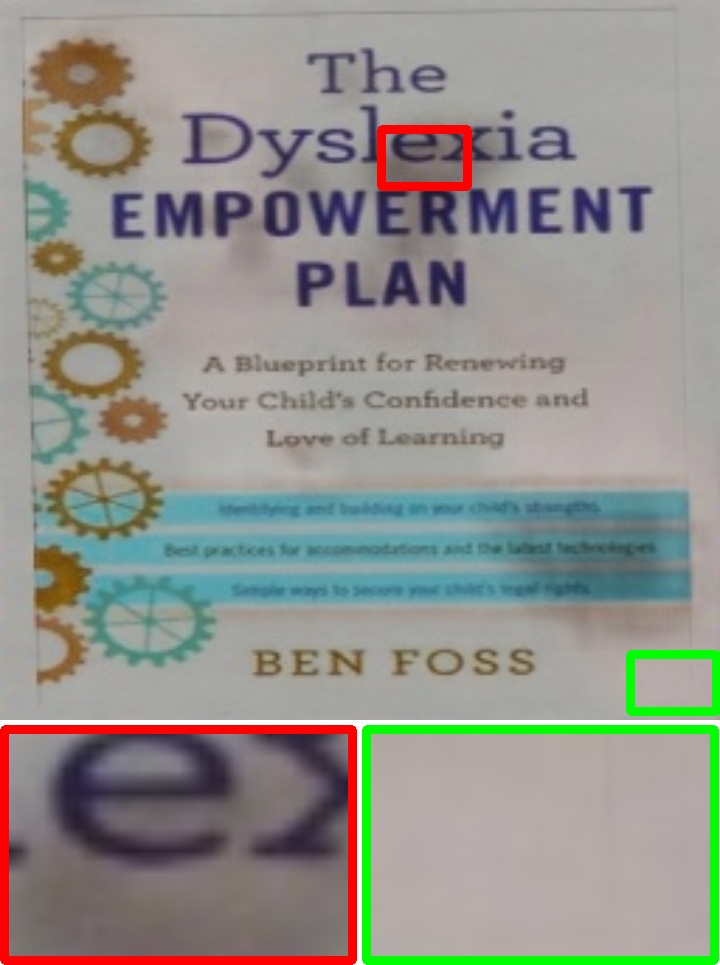}}
        \end{minipage}
        \hfill
        \begin{minipage}[b]{.16\linewidth}
            \centering
            \centerline{\includegraphics[width=\linewidth]{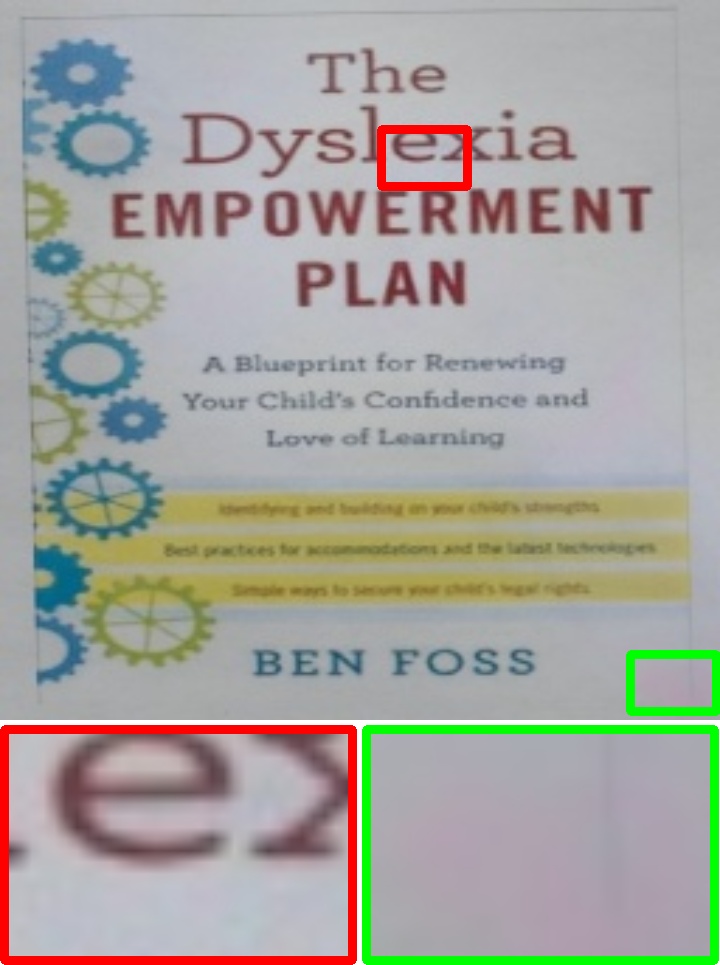}}
        \end{minipage}
        \hfill
        \begin{minipage}[b]{.16\linewidth}
            \centering
            \centerline{\includegraphics[width=\linewidth]{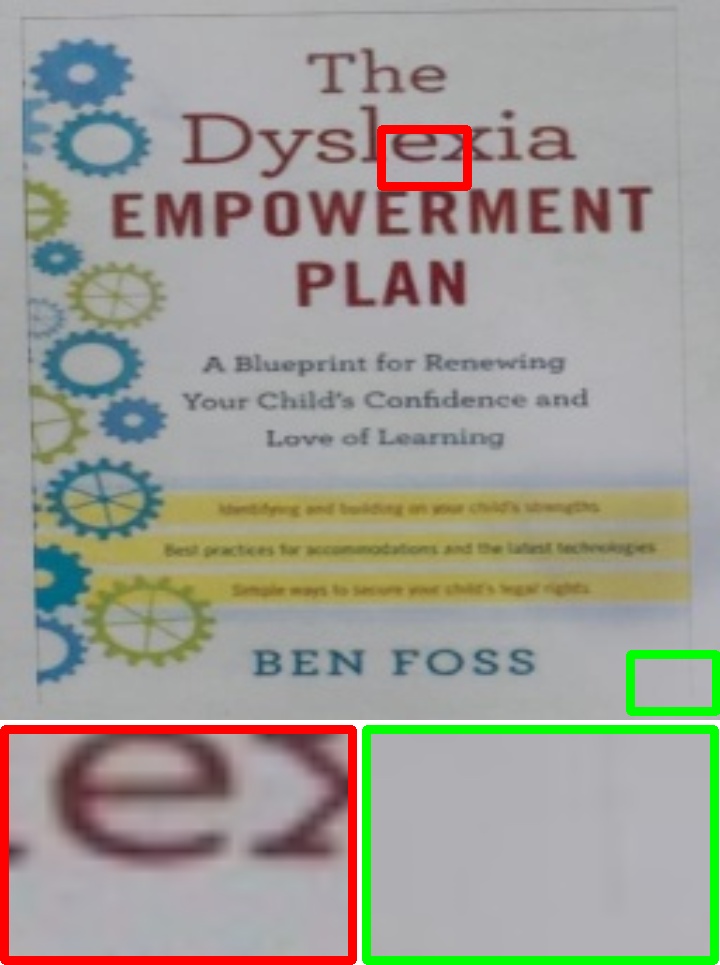}}
        \end{minipage}
        \hfill
        \begin{minipage}[b]{0.16\linewidth}
            \centering
            \centerline{\includegraphics[width=\linewidth]{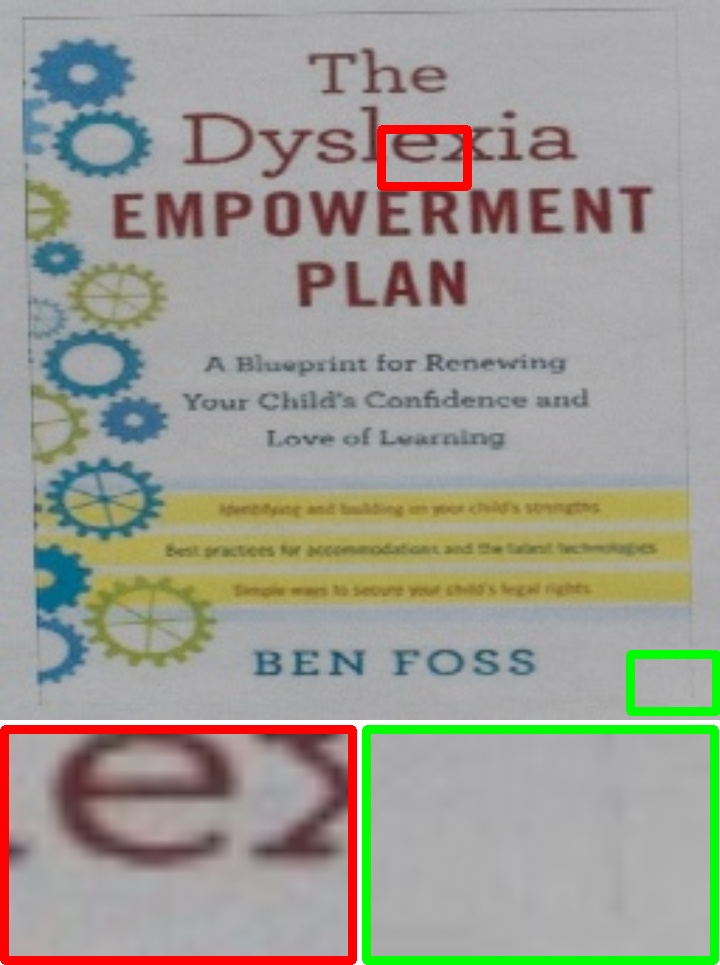}}
        \end{minipage}
    \end{minipage}

    \begin{minipage}[b]{0.9\linewidth}
        \begin{minipage}[b]{.16\linewidth}
            \centering
            \centerline{\includegraphics[width=\linewidth]{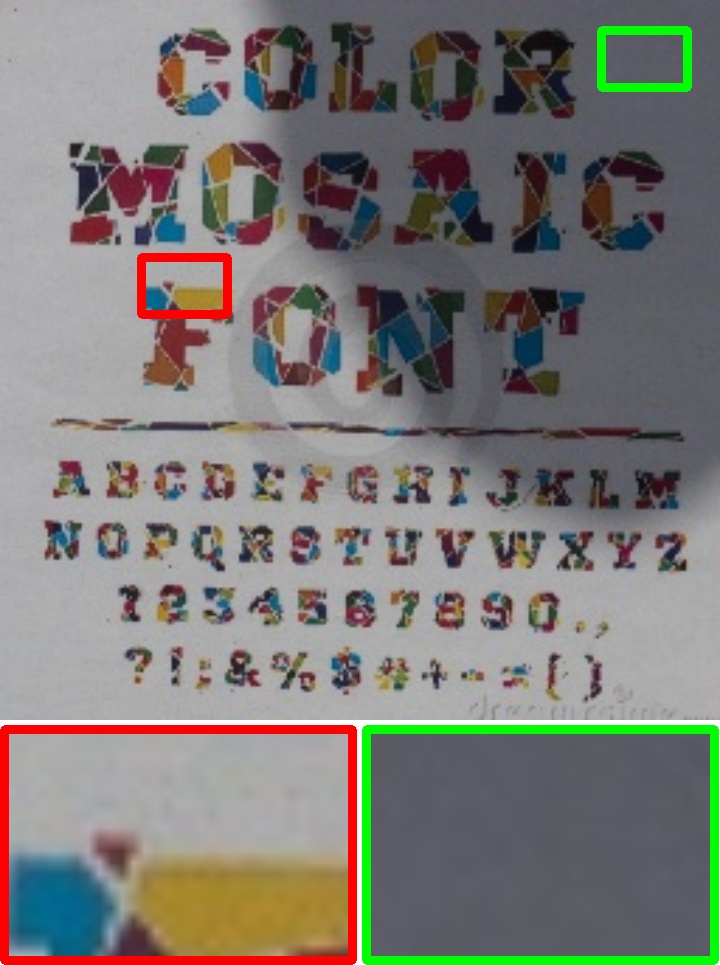}}
            \centerline{(a) Input}\medskip
        \end{minipage}
        \hfill
        \begin{minipage}[b]{.16\linewidth}
            \centering
            \centerline{\includegraphics[width=\linewidth]{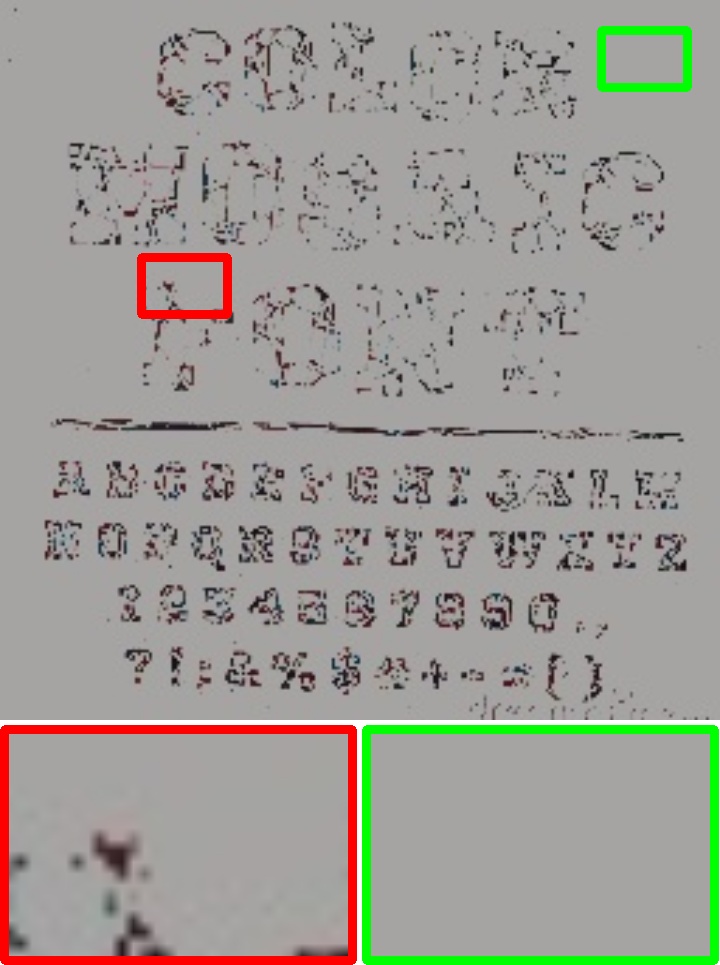}}
            \centerline{(b) Liu~\etal}\medskip
        \end{minipage}
        \hfill
        \begin{minipage}[b]{0.16\linewidth}
            \centering
            \centerline{\includegraphics[width=\linewidth]{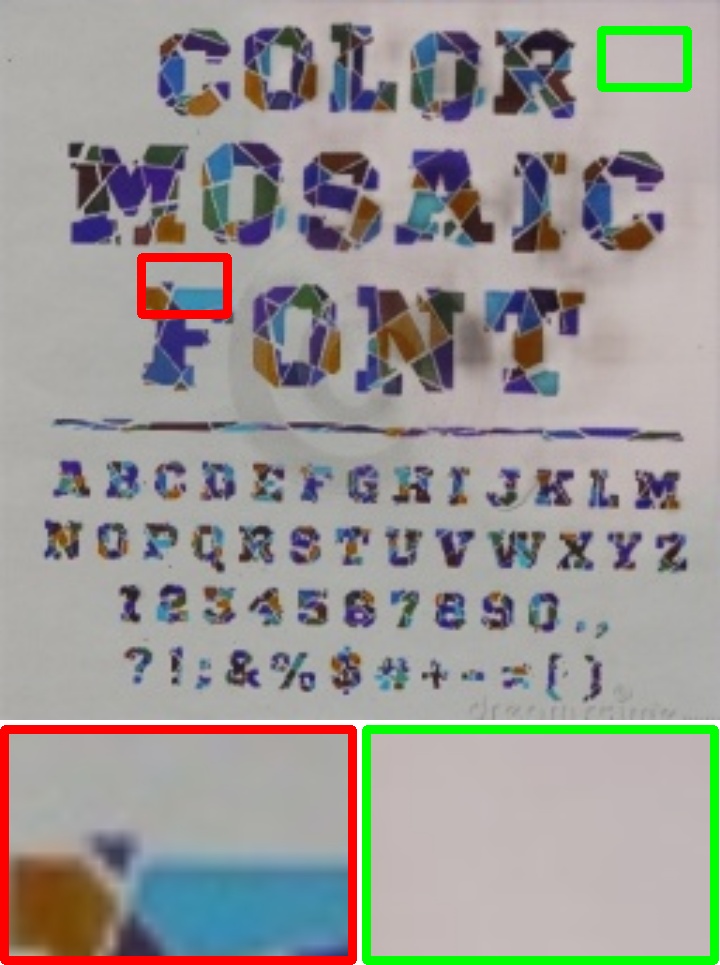}}
            \centerline{(c) BEDSR-Net}\medskip
        \end{minipage}
        \hfill
        \begin{minipage}[b]{.16\linewidth}
            \centering
            \centerline{\includegraphics[width=\linewidth]{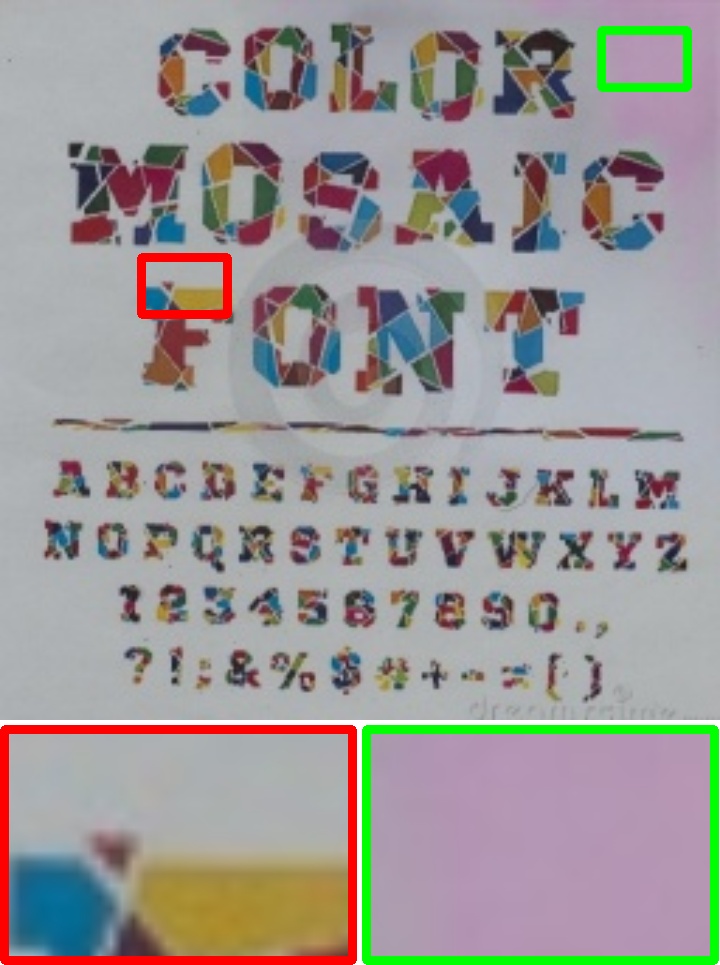}}
            \centerline{(d) BGShadowNet}\medskip
        \end{minipage}
        \hfill
        \begin{minipage}[b]{.16\linewidth}
            \centering
            \centerline{\includegraphics[width=\linewidth]{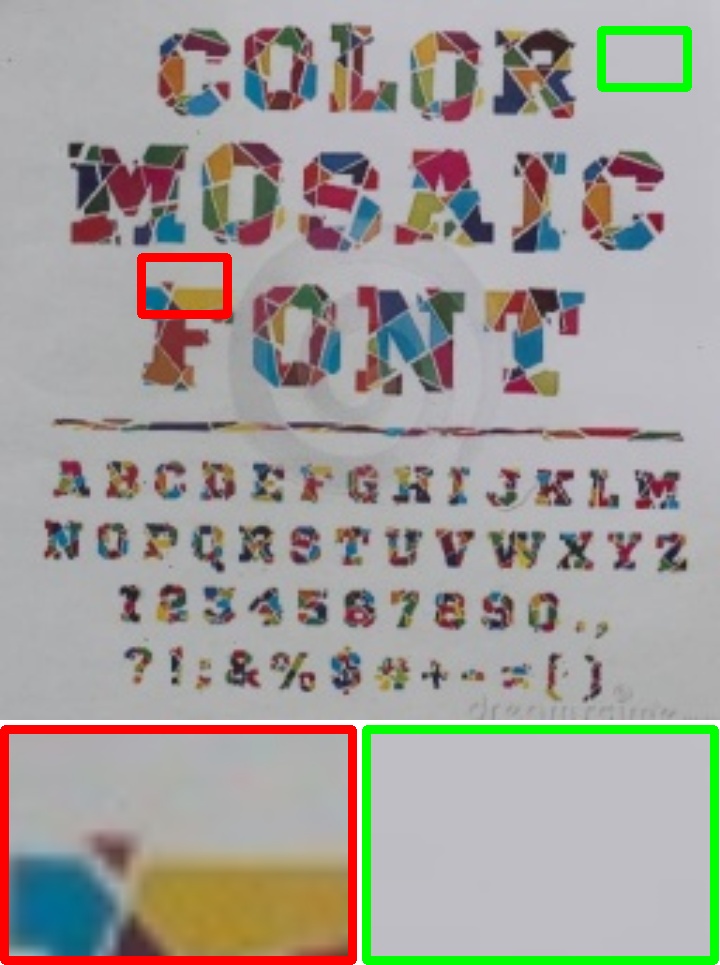}}
            \centerline{(e) Ours}\medskip
        \end{minipage}
        \hfill
        \begin{minipage}[b]{0.16\linewidth}
            \centering
            \centerline{\includegraphics[width=\linewidth]{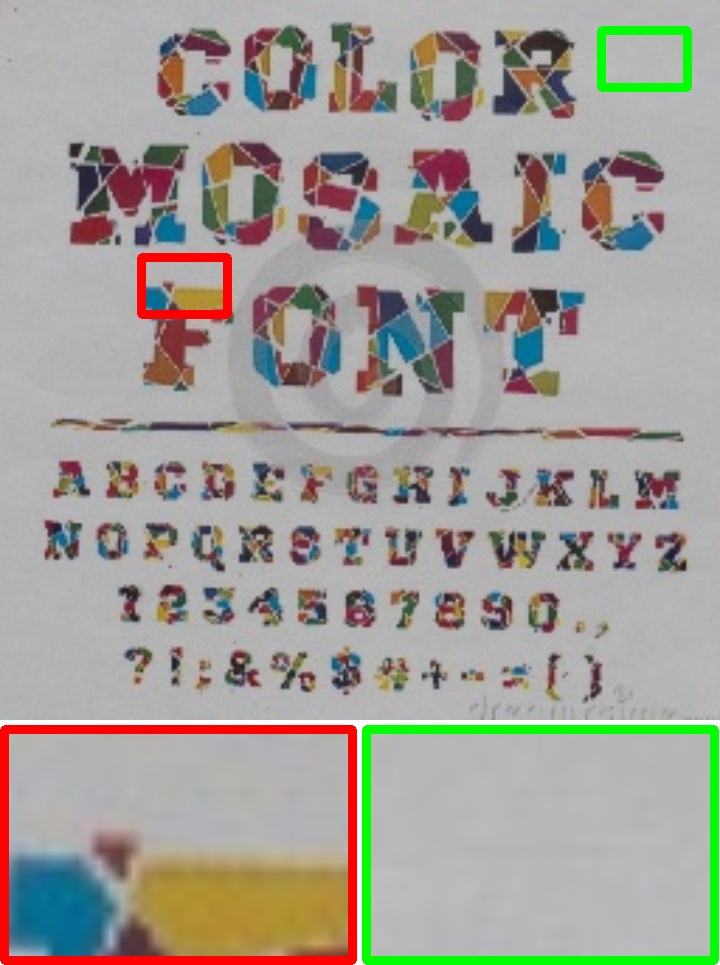}}
            \centerline{(f) Target}\medskip
        \end{minipage}
    \end{minipage}
    
    \caption{The qualitative results of comparing the methods on the RDD and Kligler dataset.} 
    \label{fig:comp}
\end{figure*}
\begin{table*}[ht]
\caption{Quantitative results of comparisons with the state-of-the-art methods. The top two results are marked in bold and underline.}
\centering
\begin{tabular}{l|ccc|ccc}
\hline
\multirow{2}{*}{Method} & \multicolumn{3}{c|}{RDD ($512\times512$)} & \multicolumn{3}{c}{Kligler ($512\times512$)} \\ \cline{2-7} 
                        & PSNR $\uparrow$   & SSIM $\uparrow$  & RMSE $\downarrow$  & PSNR $\uparrow$   & SSIM $\uparrow$  & RMSE $\downarrow$    \\ \hline
Shah~\etal~\cite{shah2018iterative}&10.50&0.79&82.36&8.64&0.70&94.83\\
Wang~\etal~\cite{wang2019effective} &13.09&0.60&68.03&16.60&0.80&39.62         \\
Jung~\etal~\cite{jung2018water}&16.32&0.84&43.95&16.84&0.87&37.49         \\
Liu~\etal~\cite{liu2023shadow}&20.38&0.85&26.30&19.37&0.80&29.40         \\
Wang~\etal~\cite{wang2020shadow}&15.85&0.83&47.07&14.55&0.78&48.99         \\
Bako~\etal~\cite{bako2017removing} &20.33&0.89&27.27&24.71&0.90&16.36         \\
Kligler~\etal\cite{kligler2018document}&16.85&0.78&37.64&21.31&0.84&22.58         \\
DeShadowNet~\cite{qu2017deshadownet}&19.37&0.87&29.38&22.19&0.86&21.08 \\
DHAN~\cite{cun2020towards}&21.81&0.91&22.23&22.95&0.9&19.07\\
LG-ShadowNet~\cite{liu2021shadow}&28.48&0.9&10.56& 24.66&0.89&15.84\\
Mask-ShadowGAN~\cite{hu2019mask}&28.22&0.91&10.97&25.53&\textbf{0.92}&14.89\\
DC-ShadowNet~\cite{jin2021dc}&\underline{28.99}&\textbf{0.93}&\textbf{9.91}&23.45&0.90&18.38         \\
BGShadowNet~\cite{zhang2023document}&10.20&0.04&80.79&10.81&0.06&74.65         \\
BEDSR-Net~\cite{lin2020bedsr} &24.41&\underline{0.92}&16.68&21.40&0.87&24.22         \\
ST-CGAN~\cite{wang2018stacked} &16.56&0.70&43.38&14.12&0.53&51.67         \\
AEFNet~\cite{fu2021auto} &27.57&\textbf{0.93}&13.82&\underline{25.60}&\textbf{0.92}&\underline{14.26}         \\  
TBRNet~\cite{tbr}&25.94&0.91&14.45&24.82&0.89&15.85\\
DMTN~\cite{dmtn}&25.32&0.89&15.84&24.35&0.89&16.06\\\hline
Ours& \textbf{29.46}   & \underline{0.92}   & \textbf{8.9}   & \textbf{26.36}    & \underline{0.90}    & \textbf{13.17}   \\ \hline
\end{tabular}
\label{table:comp}
\end{table*}
\section{Experiments}
\subsection{Datasets and Metrics}
In the context of our investigative efforts, we have undertaken a comprehensive and rigorous comparative analysis aimed at providing an in-depth examination of the efficacy of a range of different methodologies. Our evaluation process has been conducted with a high degree of precision and attention to detail, utilizing two datasets that are openly available for public use. In order to preserve uniformity throughout our analytical procedures and to guard against any potential biases that might arise from discrepancies in image resolution, we have implemented a protocol whereby all images under consideration have been resized to conform to a standard resolution of $512\times512$ pixels. This step ensures that the comparative results reflect the performance of the methodologies rather than the influence of varying image sizes.

For our research, the datasets we have chosen encompass a wide array of complexities and features that reflect the types of challenges frequently encountered in practical environments. Each dataset is distinguished by its comprehensive coverage of various scenarios and has established a reputation for consistency and dependability in the academic community. These collections of data are not only representative of the real-world conditions but have also been extensively utilized as reliable benchmarks for assessing the performance of document analysis methods. To offer a deeper understanding of the datasets employed in our study, we present a thorough overview of their compositions:

(1) RDD~\cite{zhang2023document}: 
The dataset in question has been meticulously assembled with a focus on document-based backgrounds, encompassing an array of items including printed papers, bound books, and various promotional materials. The creators of the dataset initiated the process by capturing images that feature shadows cast by objects obstructing a light source. Subsequently, they produced a set of equivalent images from which the occluding entities were eliminated, thereby creating pairs of images, one shadowed and one without shadows. The resulting Resource for Document De-shadowing (RDD) is composed of a total of 4,916 matched pairs. This collection is stratified into two subsets: one for the purpose of training, containing 4,371 image pairs, and a smaller one comprising 545 pairs reserved for the testing phase. Notably, RDD stands out as the inaugural dataset of substantial size dedicated exclusively to the task of shadow removal in the context of real-world documents.

(2) Kligler~\cite{kligler2018document}: 
In their research paper, the authors have compiled and carefully selected a dataset drawn from three well-known existing datasets. The final combined dataset that emerged from this selection process includes a total of 300 image pairs. These images have been specifically chosen and set aside for the purpose of conducting evaluations.

In order to thoroughly assess the visual quality delivered by different methods, we adopt a quantitative evaluation framework. We measure the performance using several objective metrics. The Peak Signal-to-Noise Ratio (PSNR) is utilized to evaluate the quality of image reconstruction, reflecting how much the processed image deviates from the original. We also incorporate the Structural Similarity Index (SSIM) to gauge the visual impact of the processed images, as it is designed to measure the perceived quality and visual similarity to the original image. Additionally, we use the Root Mean Square Error (RMSE), which serves as an estimator to calculate the variance between the predicted values by the model and the actual observed values. Together, these metrics offer a multifaceted view of the efficacy of the methods, allowing us to compare image integrity and the extent of errors systematically.

\subsection{Experiment Settings}
In our work, we operationalize the model within the PyTorch computational framework, deploying it on an NVIDIA RTX 2080Ti graphics processing unit. To optimize the model, we adopt the standard settings of the Adam optimization algorithm. The batch size is set to 1 and the learning rate is set to $1e-4$. To enhance the robustness of our model and to ensure its generalizability, we incorporate a series of data augmentation techniques. These augmentations encompassed a diverse set of transformations such as random cropping of images, resizing them, flipping them horizontally or vertically, and employing the mixup strategy to create composite images from the original data.


\begin{table}[ht]
\centering
\caption{The experimental results of ablation study.}
\begin{tabular}{ccc|ccc}
\hline
\multicolumn{3}{c|}{Module}                                                                                                             & \multicolumn{3}{c}{RDD ($512\times512$)} \\ \hline
\multicolumn{1}{c}{STD}                           & \multicolumn{1}{c}{EBO}                           & CDGF                          & PSNR $\uparrow$   & SSIM $\uparrow$  & RMSE $\downarrow$   \\ \hline
\multicolumn{1}{c}{\XSolidBold}    & \multicolumn{1}{c}{\CheckmarkBold} & \CheckmarkBold & 26.92  & 0.91  & 12.1   \\
\multicolumn{1}{c}{\CheckmarkBold} & \multicolumn{1}{c}{\XSolidBold}    & \CheckmarkBold & 25.39  & 0.91  & 14.21  \\
\multicolumn{1}{c}{\CheckmarkBold} & \multicolumn{1}{c}{\CheckmarkBold} & \XSolidBold    & 27.87  & 0.92  & 10.84  \\ \hline
\multicolumn{1}{c}{\CheckmarkBold} & \multicolumn{1}{c}{\CheckmarkBold} & \CheckmarkBold & \textbf{29.46}  & \textbf{0.92}  & \textbf{8.9}    \\ \hline
\end{tabular}
\label{table:ab}
\end{table}

\begin{figure*}[ht]
    \centering
    \begin{minipage}[b]{\linewidth}
        \begin{minipage}[b]{.16\linewidth}
            \centering
            \centerline{\includegraphics[width=\linewidth]{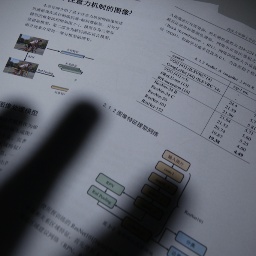}}
            \centerline{(a) Input}\medskip
        \end{minipage}
        \hfill
        \begin{minipage}[b]{.16\linewidth}
            \centering
            \centerline{\includegraphics[width=\linewidth]{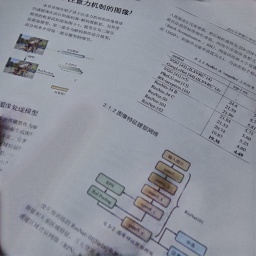}}
            \centerline{(b) NO EBO}\medskip
        \end{minipage}
        \hfill
        \begin{minipage}[b]{.16\linewidth}
            \centering
            \centerline{\includegraphics[width=\linewidth]{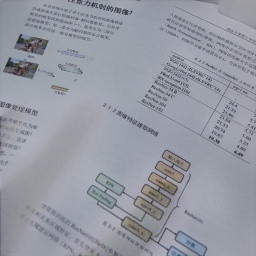}}
            \centerline{(c) NO STD}\medskip
        \end{minipage}
        \begin{minipage}[b]{.16\linewidth}
            \centering
            \centerline{\includegraphics[width=\linewidth]{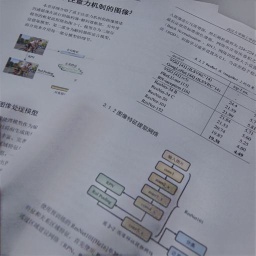}}
            \centerline{(d) NO CDGF}\medskip
        \end{minipage}
        \hfill
        \begin{minipage}[b]{.16\linewidth}
            \centering
            \centerline{\includegraphics[width=\linewidth]{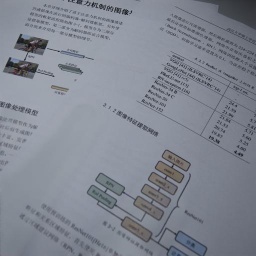}}
            \centerline{(e) Ours}\medskip
        \end{minipage}
        \hfill
        \begin{minipage}[b]{.16\linewidth}
            \centering
            \centerline{\includegraphics[width=\linewidth]{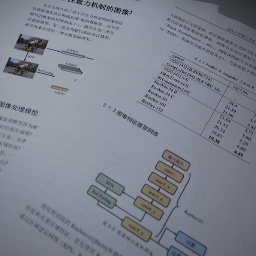}}
            \centerline{(f) Target}\medskip
        \end{minipage}
    \end{minipage}
    \caption{The visual results of the input document shadow image (a), along with the results of our model without EBO (b), without STD (c) , without CDGF (d), our full model (e) and the target (f). All the elements within our model play a crucial role in its success, and we found that neglecting even a single component led to a marked decline in the performance indicators we tracked. This outcome highlights the necessity of employing a holistic strategy to attain the superior outcomes our method has shown.
    } 
    \label{fig:ablation}
\end{figure*}

\subsection{Comparisons with State-of-the-Arts}
The outcomes of our empirical evaluation on both the RDD and Kligler datasets are systematically tabulated in Table~\ref{table:comp}. This tabulation illustrates that our ShaDocFormer method outstrips the performance of existing approaches when measured against an array of established benchmarks. Conventional strategies tend to depend on heuristics that are tailored to the peculiarities of particular datasets, which can lead to a lack of uniformity when applied to varying data sources. On the other hand, methods rooted in machine learning demonstrate robustness, consistently delivering high-caliber performance irrespective of the dataset's characteristics. Nevertheless, the prowess of these learning-based approaches may be somewhat hampered by limitations such as the finite volume of the training data and the potential for imprecise delineation of shadow regions and background estimation.

Fig.~\ref{fig:comp} provides a visual exposition of the qualitative outcomes, reinforcing the preeminence of our proposed method. In stark contrast to alternative techniques that frequently yield noticeable distortion and discoloration, our ShaDocFormer distinguishes itself by adeptly eradicating shadows. It does so while faithfully upholding the integrity of the original image's hues, culminating in an output that bears the closest resemblance to the authentic target image. This fidelity is not merely anecdotal but is underpinned by our method's innate capacity to discern and preserve the underlying color distribution of the document, achieving the highest resemblance to the target image.

\subsection{Ablation Studies}
To thoroughly assess the effectiveness of the ShaDocFormer, we perform a sequence of ablation studies. These studies are designed to isolate and identify the contribution of individual components within our method, with the results comprehensively presented in Table~\ref{table:ab}. The data we've gathered strongly supports the conclusion that the complete ShaDocFormer method exhibits significantly improved performance when contrasted with its various partial variants. Through our analytical process, we have determined that key elements such as the Shadow-attentive Threshold Detector, the Encoder Block with its Aggregation feature, and the Convolutional Depthwise Grouped Fusion Net are integral to our method's success. Fig.~\ref{fig:ablation} (b), (c) and (d) respectively show the performance of our model without EBO, STD and CDGF. It is not difficult to find that it is consistent with our full model (e ), their performance is much inferior. The performance of the full model is closest to target (f). Theredore, each of these components contributes substantially to the method's effectiveness, and omitting any one of them leads to a marked degradation in the performance metrics that we have measured. This evidence points to the importance of the integrated approach in achieving the high-quality results demonstrated by our method.

\section{Conclusion}
In this paper, we present ShaDocFormer, a state-of-the-art structure that fuses the reliable principles of traditional algorithms with the progressive power of deep learning. This innovative approach harnesses a Transformer-based architecture, deliberately crafted to tackle the challenge of shadow removal in document images. Our extensive experimental evaluation, which draws upon the data from both the RDD and Kligler collections, reveals that our model achieves a substantial uptick in performance, outstripping the current advanced methods designed for enhancing document images plagued by shadows. The model's uniform superiority in various evaluative scenarios suggests its capacity to establish a novel benchmark within the document image processing landscape, heralding a new era of methodological advancements.

\section*{Acknowledgment}
This work was supported in part by the Science and Technology Development Fund, Macau SAR, under Grant 0087/2020/A2 and Grant 0141/2023/RIA2.

\bibliographystyle{IEEEtran}
\bibliography{ref}

\end{document}